\documentclass[10pt,twocolumn,letterpaper]{article}

\usepackage[pagenumbers]{iccv} 
\usepackage{multirow}
\usepackage{float}
\usepackage{makecell}
\usepackage{arydshln}
\usepackage{graphicx}
\usepackage{array}
\usepackage{adjustbox}

\usepackage[accsupp]{axessibility}  
\usepackage{bbding}

\usepackage{amsmath}

\definecolor{iccvblue}{rgb}{0.21,0.49,0.74}
\usepackage[pagebackref,breaklinks,colorlinks,allcolors=iccvblue]{hyperref}

\def\paperID{4027} 
\def\confName{ICCV}
\def\confYear{2025}

\title{Learning Streaming Video Representation via Multitask Training}

\author{Yibin Yan$^{1,2*}$ 
, Jilan Xu$^{3,4*}$ 
, Shangzhe Di$^{1}$ 
, Yikun Liu$^{1}$  
, Yudi Shi$^{1}$, \\  
 Qirui Chen$^{1}$  
, Zeqian Li$^{1}$ 
, Yifei Huang$^{4}$  
, Weidi Xie$^{1}$ \\[3pt]
$^{1}$School of Artificial Intelligence, Shanghai Jiao Tong University \hspace{0.5cm} \\
$^{2}$Shanghai Innovation Institute\hspace{0.5cm} 
$^{3}$Fudan University\hspace{0.5cm} 
$^{4}$Shanghai AI Laboratory\hspace{0.5cm}
} 
\makeatletter
\def\hlinewd#1{%
\noalign{\ifnum0=`}\fi\hrule \@height #1 %
\futurelet\reserved@a\@xhline}
\makeatother

\begin{document}
\twocolumn[{
\renewcommand\twocolumn[1][]{#1}
\maketitle
\begin{center}
    \captionsetup{type=figure}
    \includegraphics[width=1.0\textwidth]{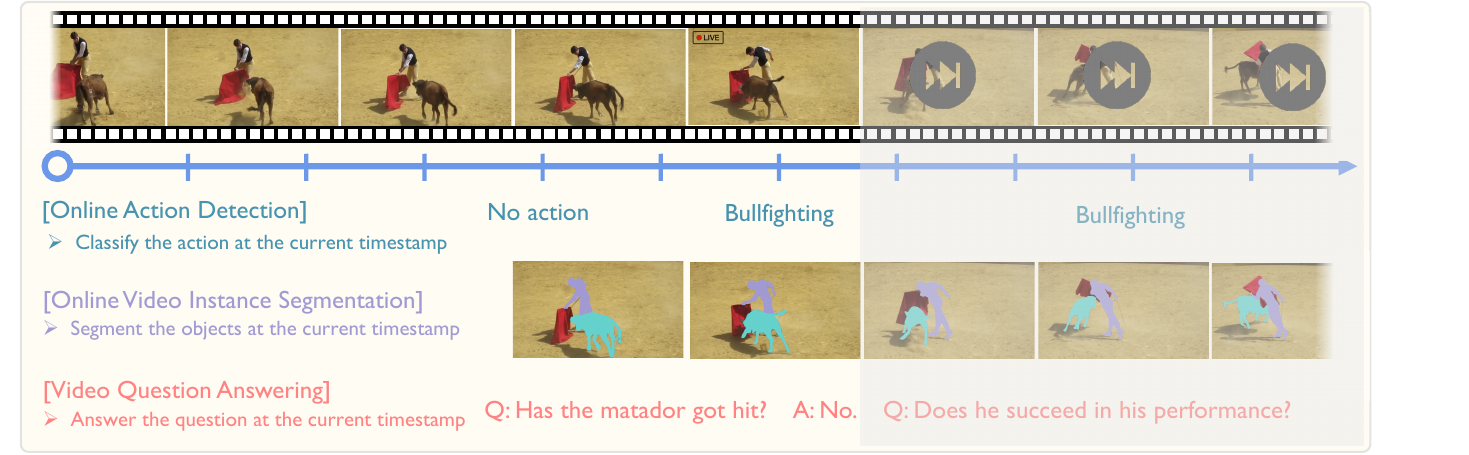}
    \caption{\textbf{StreamFormer} learns streaming video representations of various granularities through multitask training, making it applicable for diverse downstream tasks such as Online Action Detection, Online Video Instance Segmentation and Video Question Answering.}
    \label{fig:teaser}
\end{center}
}]
\renewcommand{\thefootnote}{\fnsymbol{footnote}} 
\footnotetext[1]{Equal Contribution.} 
\renewcommand*{\thefootnote}{\arabic{footnote}}
\footnotetext[1]{Project Page: \href{https://go2heart.github.io/streamformer}{https://go2heart.github.io/streamformer}}

\begin{abstract}
Understanding continuous video streams plays a fundamental role in real-time applications, including embodied AI and autonomous driving. 
Unlike offline video processing, streaming video understanding requires the ability to process video streams frame by frame, 
preserve historical information, and make low-latency decisions.
To address these challenges, our main contributions are three-fold.  
(i) we develop a novel streaming video backbone, termed as \textbf{StreamFormer}, by incorporating causal temporal attention into a pre-trained vision transformer. This enables efficient streaming video processing while maintaining image representation capability.  
(ii) to train StreamFormer, we propose to unify diverse spatial-temporal video understanding tasks within a multitask visual-language alignment framework. Hence, StreamFormer learns global semantics, temporal dynamics, and fine-grained spatial relationships simultaneously. 
(iii) we conduct extensive experiments for online action detection, online video instance segmentation, and video question answering. StreamFormer achieves competitive performance while maintaining efficiency, demonstrating its potential for real-time applications.
\end{abstract}   
\section{Introduction}

The growing demand for real-time applications, such as robotics~\cite{rt1,radosavovic2023real,driess2023palm}, autonomous driving~\cite{kiran2021deep,chitta2022transfuser,pan2024vlp}, and interactive systems~\cite{egoexo4d,vinci,streamchat}, 
has driven a shift in video understanding research from traditional offline analysis~\cite{feichtenhofer2019slowfast,timesformer,internvideo,umt} to online streaming paradigms~\cite{videollm_online,flash_vstream,rekv,streaminglong,videochat-online}. 
While offline video understanding focuses on analyzing pre-recorded videos with full temporal context, streaming video understanding addresses the challenge of dynamically processing continuous video streams in real time. 
As illustrated in Figure~\ref{fig:teaser}, 
models in this setting are required to sequentially process incoming frames, deliver timely predictions, and operate without access to future frames or the ability to frequently revisit distant past frames~\cite{zhou2024streaming,testra,lstr,buch2017sst}. This transition reflects the growing need for adaptive systems that prioritize latency, efficiency, and scalability in real-world scenarios.

Comparing to traditional offline approaches, 
streaming video understanding poses several unique challenges: 
{\em First}, efficient frame-by-frame processing is crucial, 
as incoming visual signals must be handled promptly and continuously. Unlike existing offline video backbones~\cite{feichtenhofer2019slowfast,vivit,internvideo}, 
which typically process clips as their basic unit, 
streaming models must operate seamlessly at a per-frame basis. 
{\em Second}, long-term context preservation is essential, requiring models to effectively capture temporal dependencies across frames under strict memory and computational constraints. 
{\em Third}, low-latency decision-making is indispensable, ensuring predictions or actions to be promptly delivered for real-time scenarios. Addressing these challenges necessitates specialized backbone architectures specifically designed to balance efficiency, memory management, and accuracy.

Recent works adopt pre-trained image encoders to extract frame features~\cite{zhang2024llavanextvideo,qwen2_vl}, which are subsequently processed auto-regressively by large language models (LLMs). Although this strategy aligns well with streaming video requirements, the reliance on aggregating coarse historical information within LLMs often leads to the loss of important fine-grained spatial and temporal details. To address these limitations, we propose training a streaming video backbone capable of supporting diverse spatiotemporal tasks, ranging from global scene understanding to fine-grained spatiotemporal event recognition. 
The proposed model, termed as \textbf{StreamFormer}, 
adapts Vision Transformer architectures pre-trained on images (\emph{e.g.}, SigLIP~\cite{siglip}) by introducing causal attention along the temporal dimension at the spatial-temporal feature extraction stage. This approach allows the backbone to explicitly model temporal dependencies at early vision encoding stages, resulting in richer and more temporally coherent feature representations.

While training the streaming video backbone, 
we leverage various datasets that are manually annotated by the vision community~\cite{k400,ssv2,msrvtt,anet,didemo,charades,ytvis-2019}, 
unifying various spatial-temporal video understanding tasks into a coherent multi-task learning framework. 
In particular, we combine three complementary learning objectives: 
(i) video-level classification for global semantics; 
(ii) per-frame supervision for temporal understanding, and 
(iii) pixel-wise segmentation for fine-grained spatial comprehension. By leveraging shared textual embeddings from SigLIP, 
our framework supports both closed-set and open-set tasks. 
This text-visual alignment mechanism facilitates knowledge transfer across tasks with varying granularities, enabling compatibility with diverse label spaces. 
In contrast to existing video backbone training paradigms~\cite{internvideo,videoprism,lin2022frozen} that predominantly rely on global contrastive learning using massive video-text datasets~\cite{internvid,webvid,howto100m}, our joint optimization of diverse objectives explicitly ensures capturing detailed temporal motion cues and spatial precision within streaming contexts.

In summary, our contributions are threefold:
(i) we propose \textbf{StreamFormer}, which effectively converts pre-trained image backbones into streaming video models via temporal causal attention and spatial low-rank adaptations. Our approach achieves efficient streaming video processing without sacrificing spatial representational quality; 
(ii) we introduce a hierarchical multitask learning paradigm, systematically unifying global-level, temporal-level, and spatial-level learning objectives. This paradigm enables our model to simultaneously comprehend global semantics, temporal dynamics, and fine-grained spatial relationships within streaming video;
(iii) we empirically demonstrate the effectiveness and efficiency of our method on diverse downstream tasks, including online action detection, online video instance segmentation, and video question answering. 
Our approach achieves promising performance across these tasks, effectively balancing accuracy and computational efficiency in streaming scenarios.

\section{Related Work}

\noindent \textbf{Video Backbone.}
{Early video understanding methods employ 3D convolutional neural networks~\cite{k400,tran2015learning,feichtenhofer2017spatiotemporal} or factorized convolutions~\cite{wang2016temporal,tran2018closer,tran2019video} to learn spatial-temporal features. Recent works have adopted Vision Transformers~\cite{vit} to capture long-range dependencies in videos by incorporating global spatiotemporal self-attention~\cite{vivit}, local self-attention~\cite{videoswin}, local-global self-attention~\cite{uniformer}, divided space-time attention~\cite{timesformer}, and multi-scale attention mechanisms~\cite{mvit,mvitv2}. 
Despite the progress in offline video tasks, the use of bi-directional attention in visual backbones makes them less suitable for real-time video tasks. 
In NLP, LLMs typically adopt a decoder-only architecture~\cite{llama,llama2,deepseekv3,qwen}, where causal attention enables efficient autoregressive generation during inference.
Currently, autoregressive modeling in vision primarily focuses on training backbones for image or video generation~\cite{van2016conditional,parmar2018image,bai2024sequential,sun2024autoregressive,tian2025visual}. 
For visual understanding tasks, iGPT~\cite{chen2020generative} learns visual representations by predicting next pixels, while AIM~\cite{el2024scalable} explores the scaling laws for autoregressive image pre-training. Toto~\cite{toto} trains a causal video transformer to predict the next visual patch. 
Unlike these works, we target designing StreamFormer with causal temporal attention to address various video understanding tasks.
}

\vspace{3pt} \noindent \textbf{Video Language Pre-training.}
Building on the success of image-text pretraining~\cite{clip,siglip,mae}, recent video-language models~\cite{internvideo,internvideo2,videoprism,videococa,lavila,egoinstructor,vatt} learn generalizable representations by training on massive video-text pairs~\cite{webvid,internvid,yttemporal}. 
UMT~\cite{umt} and InternVideo2~\cite{internvideo2} combine feature distilling, masked video modeling and video-text contrastive learning, to learn unimodal and multimodal representations. 
VideoPrism~\cite{videoprism} integrates global-local distillation and token shuffling to enhance video representation learning. 
Despite strong performance, these methods typically require extensive training on million-scale video-text pairs. 
In contrast, we explore leveraging the rich, fine-grained human annotations available in existing video datasets for video language training by adopting a multitask learning paradigm, achieving higher data efficiency.

\vspace{3pt} \noindent \textbf{Streaming Video Understanding.} 
Streaming video understanding tasks necessitate the processing of video inputs on a frame-by-frame basis and demand timely reactions~\cite{orthogonal,vinci}. 
Existing streaming video approaches mainly target at online action detection~\cite{trn,lstr,gatehub,mat,oaoad}, online video instance segmentation~\cite{idol,dvis++,ctvis} and dense video captioning~\cite{zhou2024streaming}. Leveraging the autoregressive paradigm in LLMs, recent approaches extract frame-wise features using pre-trained image encoders and pass them to LLMs for online question answering~\cite{rekv,flash_vstream,videochat-online}. 
However, aggregating coarse frame features often results in the loss of fine-grained spatial and temporal details. 
In contrast, StreamFormer addresses causal temporal interaction in the video backbone to preserve spatial-temporal details.

\vspace{3pt} \noindent \textbf{Multitask Learning.}
Multitask models provide different outputs for a given input, simultaneously supporting various tasks~\cite{caruana1997multitask,ruder2017overview,zamir2018taskonomy,eigen2015predicting,misra2016cross,kendall2018multi,videollm}.
Motivated by the success of unifying diverse tasks as sequence-to-sequence modeling in LLMs~\cite{raffel2020exploring,brown2020language,chowdhery2023palm}, recent works either train a unified multimodal and multitask transformer model~\cite{anygpt,ofa,unifiedio,chameleon,uniperceiver,pix2seq}, or leverage LLM as a bridge to combine task-specific heads~\cite{visionllm,visionllm2,tpo}. 
Inspired by these works, we train our streaming video backbone in a unified multitask visual-language alignment framework. 

\begin{figure*}[ht]
    \centering
    \includegraphics[width=1.0\textwidth]{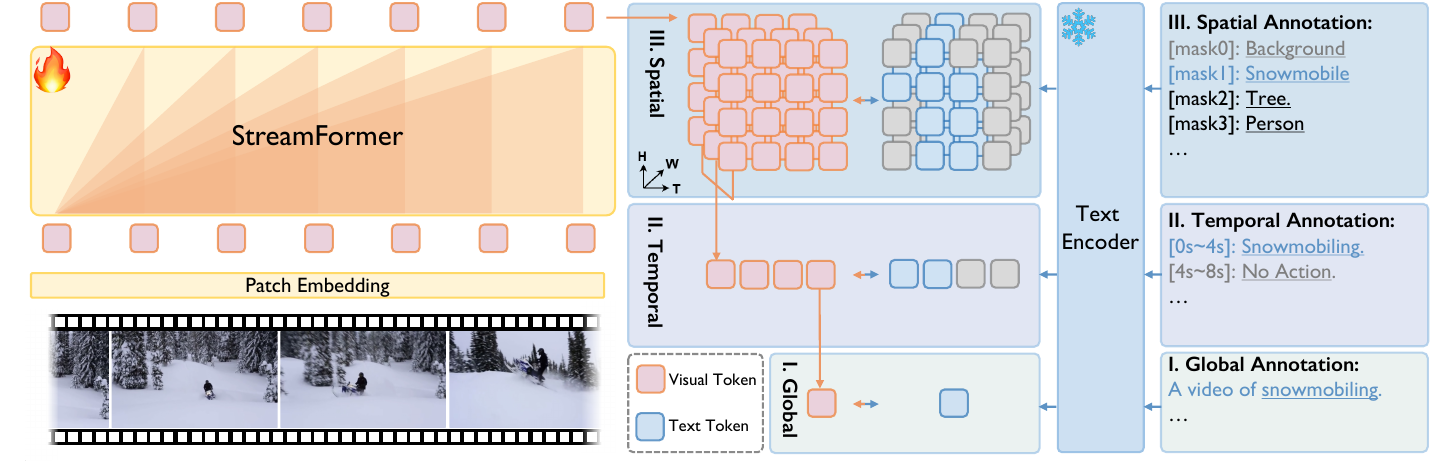}
    \vspace{-0.6cm}
    \caption{\textbf{Overall framework of StreamFormer.} Our StreamFormer is trained under a unified visual-language alignment framework, enabling simultaneous understanding of global semantics, temporal dynamics, and fine-grained spatial relationships. Each level utilizes features of different granularities: (i) last frame for the global level, (ii) per frame for the temporal level, (iii) and per frame per patch feature for the spatial level.
    }
    \label{fig:arch}
    \vspace{-0.6cm}
\end{figure*}

\section{Method} 

We begin by formulating the problem in Sec.~\ref{subsec:problem_formulation}.
Next, Sec.~\ref{subsec:video_encoding} describes the streaming video encoding backbone, followed by the language encoding model in Sec.~\ref{subsec:language_encoding}. 
Finally, in Sec.~\ref{subsec:multitask}, we elaborate on training the streaming video backbone to address diverse video understanding tasks within a unified framework, as illustrated in Figure~\ref{fig:arch}.

\subsection{Problem Formulation}
\label{subsec:problem_formulation}

This paper considers the task of streaming video understanding, 
which aims to continuously process video frames by extracting spatiotemporal features in real-time, while retaining historical context and supporting diverse video understanding tasks. To achieve this, we propose to train a unified streaming video backbone capable of capturing global semantics, temporal dependencies, and fine-grained spatial details, all within a single framework.

Specifically, given a streaming video, represented as 
$\mathcal{V}^T = [I_1, I_2, \dots, I_T]$, where $T$ denotes the total number of frames in the video. 
At any time step $t \ (1 \leq t \leq T)$, 
the visual features of different granularities can be computed as:
\begin{equation}
    \{\mathbf{v}^t, \mathbf{f}^t, \mathbf{F}^t\} = f_{\text{video}}(\mathcal{V}^t),
\end{equation}
where $\mathbf{v}^t \in \mathbb{R}^{d}$, $\mathbf{f}^t \in \mathbb{R}^{t \times d}$, 
and $\mathbf{F}^t \in \mathbb{R}^{t \times h \times w \times d}$ denote the global, temporal, and spatial video features, respectively. The computation details are provided in Sec.~\ref{subsec:video_encoding}.

To unify the diverse spatiotemporal tasks in a streaming setting, 
we reformulate them as visual-language alignment. 
Given a set of $N$ tasks, denoted as $\mathcal{Q} = \{q_1, q_2, \dots, q_N\}$, 
the output for each task $q_i$ ({\em e.g.}, action recognition, temporal localization, or object segmentation) can be viewed as an alignment score $\mathcal{A}_i$ in a shared visual-language space:
\begin{equation}
\mathcal{A}_i = g_{\text{tasks}}(f_{\text{video}}(\mathcal{V}^t), f_{\text{text}}(y), q_i),
\end{equation}
where $g_{\text{tasks}}(\cdot)$ produces the video-text alignment score $\mathcal{A}_i$ given a task $q_i$, and $f_{\text{text}}(y)$ is the text representation for a given action prompt or free-form video caption $y$ (detailed in Sec.~\ref{subsec:language_encoding}).
In this way, training multiple tasks can be unified as optimizing the alignment score $\mathcal{A}$.
As far as we know, this is the first work to develop a streaming video backbone 
that supports multiple spatiotemporal tasks under a unified framework, 
making it scalable and feasible for tackling a wide range of video understanding tasks.

\vspace{2pt} \noindent \textbf{Discussion.}
Reformulating various video understanding tasks as visual-language alignment offers several key advantages: (i) it eliminates the need for task-specific architectures by unifying all tasks as interactions between visual and textual features, enabling a streamlined and consistent framework;
(ii) it facilitates multitask learning, allowing the model to generalize across task boundaries by leveraging shared spatiotemporal representations. This improves data efficiency and reduces training overhead;
(iii) the framework is inherently extensible, supporting the incorporation of new tasks by simply defining additional alignment mechanisms, without requiring modifications to the underlying video backbone. 
By leveraging this unified framework, our method provides a flexible and efficient solution for streaming video understanding across varying complexity and granularity, 
including video semantics, temporal dependencies, and fine-grained spatial reasoning.

\subsection{Video Encoding}
\label{subsec:video_encoding}

For the input video $\mathcal{V}^T\in\mathbb{R}^{T\times H\times W\times 3}$, it is first transformed into patch embeddings, with spatial and temporal position embeddings added separately, and the resulting video representation $z\in\mathbb{R}^{T\times hw \times d}$ is the input to the video backbone. 
We adopt a series of Transformer blocks that employ divided space-time attention~\cite{timesformer}, with a key modification to make the temporal attention causal. 

Specifically, while computing the visual feature for the current frame, 
the causal temporal attention operates on the {\bf same spatial positions} across {\bf all historical frames} observed so far.
The output for the $l^{\text{th}}$ layer is denoted as:
\begin{align}
    z^{l+1} &= f_{\text{FFN}} (f_{\text{space}}(f_{\text{time}}(z^l)))
\end{align}
where $f_{\text{time}}(\cdot)$, $f_{\text{space}}(\cdot)$ and $f_{\text{FFN}}(\cdot)$ refer to causal temporal attention, spatial attention, and feed-forward network, respectively. Next, we illustrate $f_{\text{time}}(\cdot)$ and $f_{\text{space}}(\cdot)$ in detail.

\vspace{3pt} \noindent\textbf{Temporal Causal Attention.}
In contrast to the conventional bi-directional temporal attention used in video understanding backbones~\cite{vivit,timesformer}, causal temporal attention models the temporal correlations between frames in a streaming setting, where frames are processed on-the-fly as they arrive, 
and each frame is restricted to only access features from preceding frames, {\em i.e.}, future frames remain inaccessible. This constraint is enforced by applying a temporal causal mask $M \in \mathbb{R}^{T \times T}$ within the attention mechanism. At the $l^{\text{th}}$ layer, the causal temporal attention is formulated as:
\vspace{-0.4em}
\begin{equation}    
\begin{aligned}
    x^l = f_{\text{time}}(z^l) &= \text{\texttt{softmax}} \left (\frac{QK^{\mathsf{T}}}{\sqrt{d}}+M \right )V + z^l, \\
    M_{ij} &=\begin{cases}
    0, & \text{if i}\geq j,  \\
    -\infty, & \text{otherwise},
\end{cases}
\end{aligned}
\end{equation}
where $Q,K$ and $V$ are linear transformations of $z$, and \texttt{softmax} is operated along the temporal dimension.

This design enables efficient processing of videos in a streaming setup, while also preserving temporal dependencies.

\vspace{3pt} \noindent\textbf{Spatial Attention.}
The spatial attention layer takes the output of the causal temporal attention layer and models the spatial relationship within each video frame. 
Following prior works~\cite{timesformer,vivit,stadapter,li2024zeroi2v}, 
we initialize the spatial attention layers from a pre-trained image Vision Transformer (\emph{e.g.}~SigLIP~\cite{siglip}). 
To preserve its strong feature representation while enabling video-specific adaptation, we employ low-rank adaptation (LoRA)~\cite{hu2022lora} strategy, which only introduces a small number of trainable parameters to modulate the pre-trained spatial weights effectively. The spatial attention is represented by:
\begin{equation}
\begin{aligned}    
    & \hat{x}^l = \text{\texttt{softmax}} \left (\frac{Q'K'^{\mathsf{T}}}{\sqrt{d}}\right )V' + x^l. \\
    & Q'=W_Q^\mathsf{T}x^l + W_{QB}^\mathsf{T}(W_{QA}^\mathsf{T}x^l)
\end{aligned}
\end{equation}
where the learnable LoRA parameters $W_{QA}\in\mathbb{R}^{d\times r}$ and $W_{QB}\in\mathbb{R}^{r\times d}$ ($r\textless d$); $K'$ and $V'$ are transformed accordingly. \texttt{softmax} is operated along the spatial dimension.

\vspace{3pt} \noindent \textbf{Outputs.}
StreamFormer consists of 12 Transformer blocks, 
and outputs features of various resolutions: 
(1) spatiotemporal video features $\mathbf{F}\in\mathbb{R}^{T\times h\times w\times d}$ by taking the last layer's hidden states, 
(2) temporal feature $\mathbf{f}\in\mathbb{R}^{T\times d}$ by applying an attention pooling layer to $\mathbf{F}$ over the spatial dimension following SigLIP~\cite{siglip},
(3) global video feature $\mathbf{v}\in\mathbb{R}^{d}$ by taking the last frame feature. 
All visual features are projected into a shared visual-text representation space using a visual projector $\phi_{\text{visual}}(\cdot)$.
To avoid notation abuse, we reuse $\mathbf{v}, \mathbf{f}$ and $\mathbf{F}$ to represent projected features.

\subsection{Language Encoding}
\label{subsec:language_encoding}

The groundtruth annotations for all considered training tasks encompass two forms: (1) \textbf{closed-set category}: an action or object label in the video ({\em e.g.}, kicking a football), and (2) \textbf{open-ended caption}: a natural language description of the video ({\em e.g.}, a man opens the window). 

As in the former case, we construct prompt templates to convert the category labels into natural language descriptions, {\em e.g.}, a video clip of \textit{kicking a football}. Details on the prompt templates can be found in the supplementary material.
In this way, we can reformulate all groundtruth annotations into free-form languages~($y$), and compute the text embeddings~($\textbf{t}\in\mathbb{R}^{d}$), 
{\em i.e.}, $\textbf{t} = f_{\text{text}}(y)$,
where $f_{\text{text}}(\cdot)$ denotes the pre-trained text encoder from SigLIP. This visual-text alignment facilitates knowledge transfer across tasks with varying granularities, enabling compatibility with diverse label spaces. 

\subsection{Spatiotemporal Multitasking}
\label{subsec:multitask}
As for training StreamFormer, we unify a wide range of spatiotemporal tasks under a cohesive multitask learning framework. Unlike existing training paradigms that primarily use global contrastive learning on large-scale video-text pairs, we exploit the publicly available datasets from the vision community and focus on learning fine-grained spatiotemporal representations that capture both temporal dynamics and spatial precision. 
Specifically, we optimize the backbone with three complementary objectives: 
(i) capturing global semantics through video-level global supervision, (ii) modeling temporal dynamics with frame-level supervision, and (iii) enabling spatial precision via pixel-wise supervision.

\vspace{3pt} \noindent\textbf{Global-level training.} 
The objective here is to learn the global semantics of a video clip, 
encompassing both the primary action and the scene. 
To achieve this, the global-level training consists of two sub-tasks: 
action recognition~(AR) and video-text retrieval~(VTR).

In action recognition, the model assigns videos to a closed set of action labels, while in video-text retrieval, the model aligns videos with free-form natural language descriptions. The language encoding procedure has been detailed in Sec.~\ref{subsec:language_encoding}.
To preserve causal temporal modeling, we use the feature of the last frame as the video representation. This design ensures that the model accumulates the temporal information from previous frames to construct a comprehensive global video representation. For consistency with SigLIP's training process, we use sigmoid loss~\cite{siglip}:
\begin{align}
\mathcal{L}_{\text{AR}}=-\frac{1}{|\mathcal{B}|} \sum_{i=1}^{|\mathcal{B}|} \sum_{c=1}^{|\mathcal{C}|} \log \frac{1}{1 + e^{y_{ic}(-\tau \mathbf{v}_i \cdot \mathbf{t}_c + b)}}
\end{align}
where $y_{ic}$ is the label for video $\mathbf{v}_i$ and text $\mathbf{t}_c$, 
which equals 1 if aligned and -1 otherwise. 
A learnable temperature $\tau$ and bias $b$ are used to alleviate initial imbalance.

For the video-text retrieval loss $\mathcal{L}_{\text{VTR}}$, 
we simply replace the prompted action label with a free-form language description, and calculate the sigmoid loss with negative pairs sampled from the mini-batch~\cite{siglip}.

\vspace{3pt} \noindent \textbf{Temporal-level training.}
The objective here is to develop fine-grained discriminative capabilities along the temporal dimension, enabling the model to perform tasks such as per-frame action understanding and perceiving events that occur across frames. 
To achieve this, we consider two sub-tasks: 
temporal action localization (TAL) and temporal video grounding (TVG).

In temporal action localization, the model assigns each frame to either one of the pre-defined actions or background, while in temporal video grounding, 
the model enables to localize video segments corresponding to free-form natural language queries. Both sub-tasks leverage the temporal structure of videos, encouraging the model to capture temporal dependencies and align them with semantic textual representations. The language encoding procedure has been detailed in Sec.~\ref{subsec:language_encoding}. 
Formally, the temporal-level loss takes temporal feature $\mathbf{f}_i=[\mathbf{f}^1_i,\mathbf{f}^2_i,...,\mathbf{f}^T_i]\in\mathbb{R}^{T\times d}$, where $\mathbf{f}^t_i \in\mathbb{R}^d$ is the frame representation at timestep $t$. For temporal action localization with action classes $\mathcal{C}$:
\begin{align}
\mathcal{L}_{\text{TAL}} = -\frac{1}{|\mathcal{B}|T} \sum_{i=1}^{|\mathcal{B}|} \sum_{t=1}^T \sum_{c=1}^{|\mathcal{C}|} \log \frac{1}{1 + e^{y_{ic}^t(-\tau \mathbf{f}_i^t \cdot \mathbf{t}_c + b)}},
\label{eq:tal}
\end{align}
where $y_{ic}^t = 1$ if frame $t$ contains action class $c$ and $-1$ otherwise.

For temporal video grounding, we also formulate it as a contrastive learning paradigm. Specifically, when a certain frame in the video falls within the interval corresponding to its paired text query, we consider the pair as a positive sample. Negative samples include: (1) the current frame paired with other language samples in the batch; (2) background frames paired with all language samples in the batch. 
We calculate $\mathcal{L}_{\text{TVG}}$ following Eq.~\ref{eq:tal}, by replacing the action label with a free-form query.

\vspace{3pt} \noindent \textbf{Spatial-level training.}
Moving to the finest granularity, 
our spatial-level pre-training tackles pixel-wise comprehension through two sub-tasks: 
{\bf video object segmentation~(VOS)} for closed-set categories and {\bf referring video object segmentation~(RVOS)} for open-set language queries.
VOS aims to predict a class label (including the background) for each pixel in the video, while RVOS needs pixel-level grounding given a natural language query. 

To maintain the consistency of multitask learning, we directly compute dot product between each patch embedding with the corresponding textual embedding to get a coarse-grained segmentation map with logits.
After up-sampling the patch-level logits to a higher resolution (\emph{e.g.} 224$\times$224), a cross-entropy loss is utilized to optimize our predictions to the annotated masks. 
Formally, the spatial-level loss takes the spatiotemporal feature $\mathbf{F}_i=[\mathbf{F}_i^1,\mathbf{F}_i^2,...,\mathbf{F}_i^{THW}]\in\mathbb{R}^{(T\times H\times W)\times d}$, where $\mathbf{F}_i^{st}\in\mathbb{R}^d$ refers to the pixel-level representation at time $t$ and spatial location $s$.
For closed-set video object segmentation with classes $\mathcal{C}$, the loss is defined as:
\begin{align}
\mathcal{L}_{\text{VOS}} = -\frac{1}{|\mathcal{B}|THW} \sum_{i=1}^{|\mathcal{B}|} \sum_{st=1}^{THW} \sum_{c=1}^{|\mathcal{C}|} \log \frac{1}{1 + e^{y_{ic}^{st}(-\tau \mathbf{F}_i^{st} \cdot \mathbf{t}_c + b)}},
\label{eq:vis}
\end{align}
where $y_{ic}^{st} = 1$ if the pixel at spatiotemporal location $st$ belongs to class $c$ and $-1$ otherwise.

For referring video object segmentation, we extend the contrastive learning formulation in temporal video grounding to the spatiotemporal domain. Specifically, if a pixel in the video belongs to the object corresponding to the paired natural language query, we consider this pixel and the text query as a positive sample pair. Negative pairs include: (1) the current foreground pixel and other text queries within the batch, and (2) background pixels and all text queries within the batch. 
Again, we compute the loss $\mathcal{L}_{\text{RVOS}}$ using the same formulation as Eq.~\ref{eq:vis}, only changing the object label to the free-form language query.

\vspace{2pt} \noindent \textbf{Training objective.}
To jointly train the StreamFormer with global-level, temporal-level, and spatial-level tasks, we define the overall objective as:
\begin{align*}
    \mathcal{L} = \underbrace{(\mathcal{L}_{\text{AR}} + \mathcal{L}_{\text{VTR}})}_{\text{global-level}} +
    \underbrace{(\mathcal{L}_{\text{TAL}} + \mathcal{L}_{\text{TVG}})}_{\text{temporal-level}} +\underbrace{(\mathcal{L}_{\text{VOS}} +\mathcal{L}_{\text{RVOS}})}_{\text{spatial-level}}
\end{align*}

At training time, we alternately sample data from each task for a forward pass, and use gradient accumulation to perform backpropagation and parameter update collectively after iterating through all tasks, avoiding potential bias towards a specific task. Joint optimization of these objectives allows our model to learn unified video representations capable of capturing global semantics, temporal dynamics and spatial details, all within a single backbone. 

\section{Experiments}
In this section, we first present  datasets used for multitask training (Sec.~\ref{exp-data}), followed by the implementation details (Sec.~\ref{exp-implement}). We present results on different downstream tasks for evaluation (Sec.~\ref{exp-downstream}) and ablation studies (Sec.~\ref{subsec:ablation}) .

\subsection{Pre-training Data}
\label{exp-data}

We leverage various publicly available video datasets for training, 
encompassing annotations across various granularities.

The datasets are categorized based on the task, including action recognition (AR)~\cite{k400,ssv2}, video-text retrieval (VTR)~\cite{msrvtt,msvd,anet,didemo,vatex,lsmdc}, temporal action localization (TAL)~\cite{anet,fineaction,hacs}, temporal video grounding (TVG)~\cite{charades,qvhighlight,tacos,anetcaptions,didemo,queryd}
, video object segmentation (VOS)~\cite{ytvis-2019,lvvis,coco}, referring video object segmentation (RVOS)~\cite{mevis,referyoutubevos}. 
The total number of video-text pairs used for pre-training is $\sim$1M.

\subsection{Implementation Details}
\label{exp-implement}

We initialize the spatial attention layers using pre-trained weights from SigLIP~\cite{siglip} with LoRA rank 32, while the temporal attention layers are randomly initialized. 
To facilitate smooth integration of temporal information, 
we incorporate a learnable \texttt{tanh} gate~(initialized to zero) in the temporal attention. This design ensures the model initially preserves SigLIP's strong spatial encoding capabilities, and gradually learns temporal dependencies. 
At training time, we perform over-sampling to balance the data scale of different tasks. In all experiments, we use the base model~(\texttt{siglip-base-patch16-224}). 
We train our model for 20 epochs, with global batch size 128, and a learning rate of $1e^{-6}$. By default, we uniformly sample 16 frames from the video as input. After training StreamFormer jointly on multiple tasks, we transfer the model to downstream video tasks by fixing the backbone and training task-specific heads.  

\subsection{Comparison on Downstream Tasks}
\label{exp-downstream}
We freeze StreamFormer and evaluate it on multiple downstream tasks, including online action detection, online video instance segmentation, and video question answering.

\subsubsection{Online Action Detection}
Given an input video stream, online action detection~(OAD) requires the model to classify each frame of the video. Notably, the model can only access the current and historical frames, without seeing the future frames.  This characteristic inherently makes OAD a suitable benchmark for evaluating causal video models. 
We choose MAT~\cite{mat} as our baseline OAD model, by replacing the TSN feature~\cite{wang2016temporal} with our StreamFormer feature.  
To ensure a fair comparison, we also incorporate optical flow features extracted using the BN-Inception model~\cite{ioffe2015batch}, which are concatenated with visual features and fed into the OAD model. The evaluation metric is per-frame \textit{mean Average Precision (mAP)} on THUMOS-14 dataset~\cite{thumos}.

As shown in Table~\ref{tab:oad}, our model achieves state-of-the-art results, demonstrating a significant improvement over the baseline SigLIP without using optical flow (58.9 vs.~68.4), and even achieves comparable performance with early approaches that use flow. 
Given the time-consuming nature of extracting optical flow~\cite{gatehub,lstr}, our approach clearly offers more efficient online inference without relying on flow, while still ensuring a high accuracy in action detection.

\begin{table}[t]

\begin{center}
\setlength{\tabcolsep}{2.95mm}
\begin{small}

\begin{tabular}{lccc}
\toprule
\multirow{2}{*}{Method} & \multirow{2}{*}{Backbone} & \multicolumn{2}{c}{THUMOS-14} \\ 
\cmidrule{3-4}
& & {w/ Flow} &{w/o Flow} \\

\midrule
TRN~\cite{trn} & ResNet50 & 62.1 & - \\
OADTR~\cite{oadtr} & ResNet50 & 65.2 & - \\
Colar~\cite{colar} & ResNet50 & 66.9 & - \\
LSTR~\cite{lstr} & ResNet50 & 69.5 & 63.5 \\
GateHUB~\cite{gatehub} & ResNet50 & 70.7 & 66.5 \\
TesTra~\cite{testra} & ResNet50 & 71.2 & -  \\

\midrule
TRN~\cite{trn} & TimeSformer & 68.5 & - \\
LSTR~\cite{lstr} & TimeSformer & 69.6 & - \\
GateHUB~\cite{gatehub} & TimeSformer & 72.5 & -\\
\midrule
MAT~\cite{mat} & ResNet50 & 71.6 & 59.4 \\
MAT~\cite{mat} & SigLIP ViT & 71.1 & 58.9 \\
MAT~\cite{mat} & \textbf{StreamFormer} & \textbf{73.9} & \textbf{68.4}  \\
\bottomrule
\end{tabular}

\vspace{-5pt}
\caption{
\textbf{Comparison on Online Action Detection.}
}
\vspace{-5pt}
\label{tab:oad}

\end{small}
\end{center}
\vskip -0.1in
\end{table}

\begin{table}[t]

\begin{center}
\setlength{\tabcolsep}{0.9mm}
\begin{small}
\begin{tabular}{lccccc}
\toprule

\multirow{2}{*}{Method}&\multirow{2}{*}{Res.}  &\multirow{2}{*}{Backbone} &\multirow{2}{*}{\makecell[c]{COCO-init. }} & \multicolumn{2}{c}{YTVIS19}\\
\cmidrule{5-6}
&&&& AP & $\text{AR}_{1} $ \\
\midrule
CrossVIS~\cite{crossvis}  & 360p & ResNet50& \Checkmark & 33.7 & 35.4 \\
MinVIS~\cite{minvis}  & 360p& ResNet50 & \Checkmark& 47.4 & 45.7 \\
IDOL~\cite{idol} & 360p& ResNet50 & \Checkmark & 49.5 & 47.7 \\
DVIS++~\cite{dvis++}  & 480p& ResNet50 & \Checkmark&  55.5 & 51.1 \\
CTVIS~\cite{ctvis} & 360p&ResNet50  & \Checkmark& 55.1 &51.9  \\
\midrule

CTVIS~\cite{ctvis}  & 224p& ResNet50& \XSolidBrush & 43.6 & 42.7 \\
CTVIS~\cite{ctvis} & 224p& SigLIP ViT & \XSolidBrush & 40.9 & 41.8 \\
CTVIS~\cite{ctvis} & 224p& \textbf{StreamFormer} & \XSolidBrush & \textbf{45.1} & \textbf{43.6} \\
\bottomrule
\end{tabular}
\vspace{-5pt}
\caption{
\textbf{Comparison on Online Video Instance Segmentation.}
}
\label{table:vis}
\vspace{-5pt}

\end{small}
\end{center}
\vskip -0.1in
\end{table}

\begin{table}[t]

\begin{center}
\setlength{\tabcolsep}{1.5mm}
\begin{small}

\begin{tabular}{lccc}
\toprule

{Method} & Res. & {VideoMME} &{MLVU} \\
\midrule

LLaMA-VID~\cite{llamavid} & $\text{224}^2$ & 25.9 & 18.1 \\
Video-LLaVA~\cite{videollava} & $\text{224}^2$ & 39.9 & 18.8 \\
Video-Chat2~\cite{videochat2} & $\text{224}^2$ & 42.3 & 44.5 \\
LLaVA-NeXT-Video~\cite{zhang2024llavanextvideo} & $\text{336}^2$  & 41.0 & 36.9 \\
VideoLLaMA2~\cite{videollama2} & $\text{336}^2$ & \textbf{47.9} & \textbf{48.4} \\
\midrule
LLaVA-Next (\textbf{StreamFormer}) & $\text{224}^2$ &45.0 & \textbf{48.4} \\
\bottomrule
\end{tabular}
\vspace{-5pt}
\caption{
\textbf{Comparison on VideoQA.} All language models are 7B. 
}
\label{tab:vqa}
\vspace{-15pt}

\end{small}
\end{center}

\end{table}

\subsubsection{Online Video Instance Segmentation}
The goal of online video instance segmentation is the simultaneous detection, segmentation, and tracking of instances in videos, in a frame-by-frame manner. With our causal temporal attention, we can evaluate our video backbone model like an image backbone. 
We choose CTVIS~\cite{ctvis} as our baseline, and keep all hyperparameters identical while changing the backbone into frozen StreamFormer. 

Due to the requirement for dense features in this task, we add ViT-Adapter~\cite{vit-adapter, dvis++} to our backbone. 
Note that we \emph{do not} initialize model parameters with pre-training on COCO~\cite{coco}, and instead directly train YoutubeVIS-2019~\cite{ytvis-2019} with COCO joint training.

We report the Average Precision (AP) and Average Recall on YoutubeVIS-2019 benchmark.

As illustrated in Table~\ref{table:vis}, StreamFormer outperforms the baseline approaches (\emph{i.e.} ResNet50 and ViT) under the same experimental setups, while approaching prior works that adopt advanced training strategies (\emph{e.g.} high-resolution training or using mask head pre-trained on COCO).  

\begin{table*}[htbp]
\begin{center}
\setlength{\tabcolsep}{2.1mm}
\begin{small}
\begin{tabular}{l ccc ccc ccc cc c}
\toprule
\multirow{3}{*}{Method} & \multirow{3}{*}{\text{Global}} & \multirow{3}{*}{\text{Temporal}} & \multirow{3}{*}{\text{Spatial}} & & \multicolumn{2}{c}{Online Action Detection} & & \multicolumn{2}{c}{Online VIS} & & \multicolumn{2}{c}{VideoQA} \\ 
\cmidrule{6-7}\cmidrule{9-10}\cmidrule{12-13}
 & & & & & \multicolumn{2}{c}{THUMOS-14} & & \multicolumn{2}{c}{YouTubeVIS-2019} & & \multirow{2}{*}{VideoMME} & \multirow{2}{*}{MLVU} \\ 
\cmidrule{6-7}\cmidrule{9-10}
  & & & && \multicolumn{1}{>{\centering}p{1.2cm}}{w/ Flow} & \multicolumn{1}{>{\centering}p{1.2cm}}{w/o Flow} && \multicolumn{1}{>{\centering}p{1cm}}{AP} & \multicolumn{1}{>{\centering}p{1cm}}{$\text{AR}_{1}$} && \\
\midrule
 
SigLIP & & & & & 71.1 & 58.9 & & 40.9 & 41.8 & & 40.8 &46.1 \\
\hdashline
StreamFormer & \Checkmark & & & & 73.1 & 65.9 & & 41.9 & 41.6 & & 43.7 & 44.6 \\
StreamFormer & & \Checkmark & & & 72.1 & 65.5 & & 40.1& 41.1 & & 44.0 &\textbf{49.6}  \\
StreamFormer & & &\Checkmark & &70.3 & 51.5& & 42.3 &41.6 & & 30.0 & 29.2\\
StreamFormer & \Checkmark& \Checkmark& & & \textbf{74.1} & \textbf{69.1} & &41.9&41.4 & & 43.1 & 49.2\\
StreamFormer & \Checkmark& \Checkmark& \Checkmark & & 73.9 & 68.4 & & \textbf{45.1} &\textbf{43.6} & & \textbf{45.0} & 48.4  \\
\bottomrule
\end{tabular}
\vspace{-5pt}
\caption{
\textbf{Ablation study for pre-training tasks.}
}
\label{tab:ablation}
\vspace{-15pt}
\end{small}
\end{center}
\vskip -0.1in
\end{table*}

\subsubsection{Video Question Answering}
Building upon the frameworks established by LLaVA~\cite{liu2023visual} and LLaVA-Next-Video~\cite{zhang2024llavanextvideo}, we propose to integrate StreamFormer with large language models for video question answering. 
For a fair comparison, we replace the CLIP in LLaVA-Next (Vicuna-7b) with StreamFormer~(frozen in all training stages), and keep the remaining architecture and training pipeline unchanged. We use the same image-text pairs in LLaVA-NeXT, and sample 80K video-text pairs from LLaVA-Video-178K~\cite{llava-video-178k} . 
We provide our full training details in the supplementary materials. 

Table~\ref{tab:vqa} presents the comparison results on VideoMME~\cite{videomme} and MLVU~\cite{MLVU}. 
StreamFormer exhibits superior performance among methods that use 224$\times$224 resolution, indicating that enhancing temporal motion and spatial detail perception in video backbone enables better VideoQA.

\subsection{Ablation Study}
\label{subsec:ablation}
In this part, we study the impact of different pre-training tasks and investigate the effects of multiple design choices. 
Apart from the task ablations~(Table~\ref{tab:ablation}), 
we conduct comparative analysis experiments using 10\% data by default.

\vspace{3pt}
\noindent\textbf{Ablation study for pre-training tasks.}
Table~\ref{tab:ablation} shows the model's downstream performance when pre-trained on different task combinations. 
Our baseline is the original image encoder of SigLIP,
\emph{i.e.}, without multitask training and temporal modeling.
We draw the following conclusions: 
(1) the baseline achieves relatively good performance on downstream tasks, highlighting the strong image representation capability of SigLIP; 
(2) the incorporation of any form of video-language pre-training, either global, temporal, or spatial, yields a notable improvement over the baseline on all downstream tasks; 
(3) both global- and temporal-level tasks yield better results across various downstream tasks, while models trained solely on spatial-level tasks perform sub-optimally in online action detection and video question answering. This may be attributed to the relatively limited amount of data available for video segmentation, leading to an insufficient understanding of verbs/actions; 
(4) by integrating tasks of varying granularities, the model achieves competitive downstream performance, proving the effectiveness of our multitask training approach.

\begin{table}[t]

\begin{center}
\begin{small}
\begin{tabular}{lccc}
\toprule
Method & No TA & Bi-directional TA & Causal TA \\

\midrule
w/o Flow & 61.7 & 62.8 & 62.5 \\
\bottomrule
\end{tabular}
\vspace{-5pt}
\caption{
\textbf{Ablation study for temporal attention.}
}
\label{tab:causal}
\end{small}
\vspace{-15pt}
\end{center}
\end{table}

\vspace{3pt}
\noindent\textbf{Effect of temporal causal modeling.}
We investigate the effect of temporal causal attention (Causal TA) by comparing it with two baselines: 
(1) an image-based model that only employs spatial attention combined with LoRA, without any temporal attention (No TA); 
(2) a model utilizes bi-directional temporal attention (Bi-directional TA). 
We report the performance of the models without optical flow to verify the temporal effect of visual features. 
As observed from Table~\ref{tab:causal}, temporal attention matters in online action detection, while bi-directional attention performs slightly better than causal attention. This is reasonable as each frame can access all frames in bi-directional attention.

\begin{figure}[t]
    \centering
    \begin{subfigure}[b]{0.49\linewidth}
        \centering
        \includegraphics[width=\linewidth]{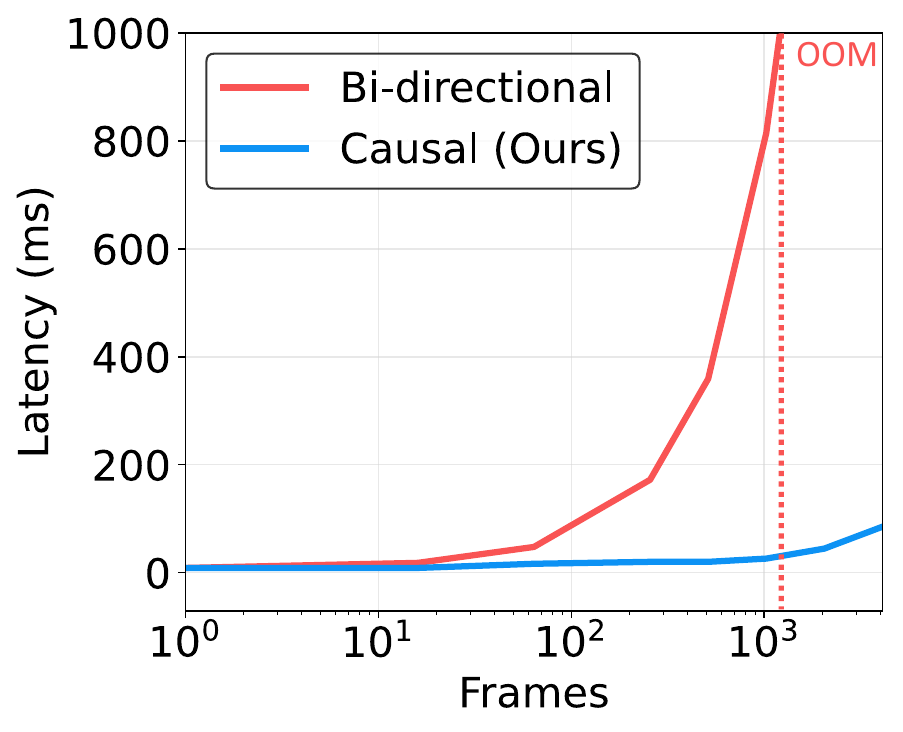}
        \label{fig:latency}
    \end{subfigure}
    \hfill
    \begin{subfigure}[b]{0.49\linewidth}
        \centering
        \includegraphics[width=\linewidth]{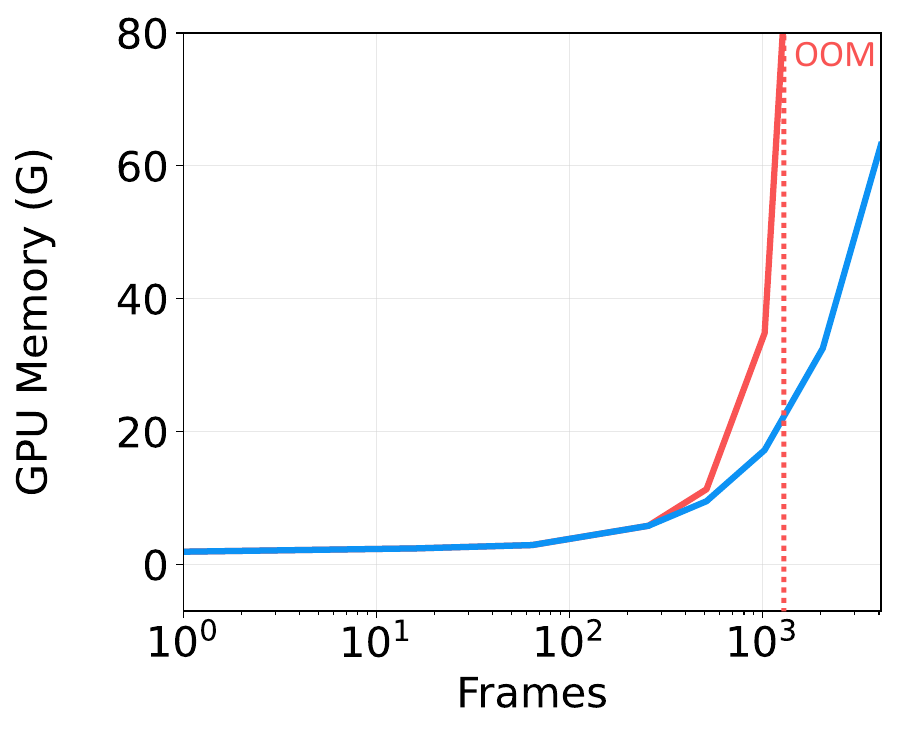}
        \label{fig:flops}
    \end{subfigure}
    \vspace{-20pt}
    \caption{
    \textbf{Computational complexity.}
    }
    \label{fig:efficiency}
    \vspace{-10pt}
\end{figure}

However, one notable advantage of causal attention over bi-directional attention is the efficiency at inference time in the streaming setting, particularly when supported by KV-Cache. 
We compare the inference cost of the backbone (\emph{i.e.}, \textit{latency} and \textit{GPU memory}) in Figure~\ref{fig:efficiency}, with input frames varying from 1 to 4096. The latency of causal attention (w/ KV-Cache) remains nearly constant as input frame increases, whereas bi-directional attention experiences a significant rise in computational cost. This highlights the scalability of StreamFormer in streaming video processing.

\vspace{2pt}
\noindent\textbf{Data efficiency.}
In this work, we leverage existing video benchmark datasets, including rich annotations of various granularities for training StreamFormer. 
In contrast, most existing video foundation models~\cite{internvideo,internvideo2,videoprism,videococa,lin2022frozen} are trained on extensive (usually $\textgreater$50M) video-text pairs~\cite{webvid,howto100m,internvid}, which is prohibitively expensive. 
For comparison, we also train our backbone on the WebVid~\cite{webvid} dataset, using the video-language contrastive loss. 

We randomly select 1M video-text pairs from WebVid-10M, 
which is comparable to the scale of our pre-training data.
As shown in Table~\ref{tab:webvid}, the models trained on WebVid-1M exhibit relatively low performance, possibly due to insufficient pre-training data for video-text contrastive learning. 
In comparison, our approach outperforms WebVid-1M model using only 0.1M pre-training data, significantly reducing the computational cost for training.  
Our method effectively leverages the existing high-quality annotated video datasets to enhance the backbone's capabilities, which suggests there is certainly a trade-off between computation and annotation cost.
\begin{table}[t]
\setlength{\tabcolsep}{1.4mm}
\begin{center}
\begin{small}
\begin{tabular}{lcccc}
\toprule
\multirow{2}{*}{Dataset} & \multicolumn{2}{c}{OAD} & \multicolumn{2}{c}{VideoQA} \\ 
\cmidrule{2-3}\cmidrule{4-5}
& {w/ Flow} &{w/o Flow}& VideoMME & MLVU \\
\midrule
WebVid-1M & 69.3 & 57.2 & 37.4 & 46.6 \\
\midrule
Ours-0.1M & 73.0 & 62.5 & 43.1 & \textbf{49.6}\\
Ours-1M & \textbf{73.9} & \textbf{68.4} & \textbf{45.0} & 48.4 \\
\bottomrule
\end{tabular}

\vspace{-5pt}
\caption{
\textbf{Comparison on data efficiency.}
}
\vspace{-0.9cm}
\label{tab:webvid}
\end{small}

\vskip -0.1in
\end{center}
\end{table}

\section{Conclusions}
In this paper, we present StreamFormer, a streaming video backbone that incorporates causal temporal attention and spatial low-rank adaptation into a pre-trained SigLIP image encoder. 
To facilitate the training of StreamFormer, we devise a multitask visual-language alignment framework that unifies diverse video understanding tasks, enabling it to learn global-level semantics, temporal-level dynamics and spatial-level perception simultaneously. 
We conduct extensive experiments on online action detection, online video instance segmentation and video question answering. 
StreamFormer exhibits promising online video understanding ability while preserving high efficiency, demonstrating its huge potential in future real-time applications.

\section*{Acknowledgement}
Weidi would like to acknowledge the funding from Scientific Research Innovation Capability Support Project for Young Faculty~(ZY-GXQNJSKYCXNLZCXM-I22).

{
    \small
    \bibliographystyle{ieeenat_fullname}
    \bibliography{main}
}
\clearpage
\setcounter{page}{1}

\onecolumn

{
\centering
\Large
\textbf{Learning Streaming Video Representation via Multitask Training}\\
\vspace{0.75em}
Supplementary Material\\
\vspace{2.0em}
}

\setcounter{section}{0}
\section{Pre-training Datasets}
We list StreamFormer's pre-training datasets in Table ~\ref{tab:dataset}. Note that for our Video Object Segmentation datasets, we utilize the VIS datasets and VIS-style produced COCO-pseudo videos for their relatively well correspondence within the objects inside the videos or pseudo videos.
\begin{table}[h]

\begin{center}

\setlength{\tabcolsep}{1mm}

\begin{tabular}{llr}
\toprule
Task & Pre-training Dataset & Scale\\

\midrule
\multicolumn{2}{l}{\small{\textit{\textcolor{gray}{Global-level}}}} \\
 \textbf{AR} & K400~\cite{k400}, SSv2~\cite{ssv2} & 400K \\
 \textbf{VTR} & \makecell[l]{ 
 MSRVTT~\cite{msrvtt}, MSVD~\cite{msvd}, ActivityNet~\cite{anet},
 DiDeMo~\cite{didemo}, LSMDC~\cite{lsmdc}, VATEX~\cite{vatex}}
 & 94K \\

 \midrule
\multicolumn{2}{l}{\small{\textit{\textcolor{gray}{Temporal-level}}}} \\
 \textbf{TAL} & ActivityNet-1.3~\cite{anet}, FineAction~\cite{fineaction}, HACS~\cite{hacs} & 180K \\

 \textbf{TVG} & \makecell[l]{ 
 CharadesSTA~\cite{charades},
 QVHighlights~\cite{qvhighlight}, TaCoS~\cite{tacos},
 ANet-Captions~\cite{anetcaptions}, DiDeMo~\cite{didemo}, QuerYD~\cite{queryd}}
 & 120K \\
 
 \midrule
\multicolumn{2}{l}{\small{\textit{\textcolor{gray}{Spatial-level}}}} \\
 \textbf{VOS} & YouTubeVIS-19~\cite{ytvis-2019}, LVVIS~\cite{lvvis}, COCO~\cite{coco} & 120K \\
 \textbf{RVOS} & MEVIS~\cite{mevis}, Refer-YouTube-VOS~\cite{referyoutubevos} & 36K \\
 \midrule
 Total & - & $\sim$1M \\
\bottomrule
\end{tabular}

\caption{
\textbf{Pretraining datasets.}
The abbreviations are defined as follows: ``AR'' for action recognition, ``VTR'' for video-text retrieval, ``TAL'' for temporal action localization, ``TVG'' for temporal video grounding, ``VOS'' for video object segmentation, and ``RVOS'' for referring video object segmentation.
}

\label{tab:dataset}

\end{center}

\end{table}

\section{Downstream Implementation Details}

\subsection{Online Action Detection}
In Online Action Detection, we adopt MAT~\cite{mat} as our baseline detection head due to its strong detection performance. Following prior works~\cite{oadtr,lstr,testra}, we extract the video frames at 24 fps, using a video chunk size of 6 (\emph{i.e.}, 4 features per second). To ensure a fair comparison, we also incorporate optical flow features extracted using the BN-Inception model~\cite{ioffe2015batch}, which are concatenated with visual features and fed into the OAD model. We train the model using Adam optimizer for 25 epochs, with a batch size of 16 and a learning rate of 7$\times 10^{-5}$. All experiments are conducted on a single NVIDIA A100 GPU.

\subsection{Online Video Instance Segmentation}
For Online Video Instance Segmentation, we use CTVIS~\cite{ctvis} as our baseline, and keep all head modules' hyper parameters the same. For our backbone, we train and test all frames resizing the short side to 224p. The batch size per device is 8 with an initial learning rate of 0.0001 and decays at 12,000 and 24,000 iterations, respectively on 4 NVIDIA H100 GPUs.

\subsection{Video Question Answering}
For Video Question Answering task, we adopt the LLaVA-NeXT-Video~\cite{zhang2024llavanextvideo} framework as our code base. To make our comparisons fair, we use the same model architecture as LLaVA-NeXT~(Vicuna-1.5-7B~\cite{vicuna}) except for our StreamFormer~(SigLIP-ViT-Base-Patch16-224~\cite{siglip}) replacing the original CLIP-Vit-Large-Patch14-336~\cite{clip}. Note that our StreamFormer is initialized from SigLIP and multi-tasking trained with SigLIP's Text Encoder, thus making it still suitable for pre-training and supervised fine-tuning on image-text data. For image datasets, we use the LAION/CC/SBU 558K~\cite{llava-1.5} for pretraining, LLaVA-NeXT-Data 779K~\cite{zhang2024llavanextvideo} for supervised fine-tuning on image-text. For Video dataset, we sample 10\% from LLaVA-Video-178K's~\cite{llava-video-178k} academic and YouTube short videos~(0 - 60s), resulting in a total 86.4K video-text samples. For model evaluation, we use the lmms-eval toolkit~\cite{lmms-eval} with 16 frames per video. We run all experiments of VideoQA on 4 NVIDIA H100 GPUs. The detailed configuration is shown in Table~\ref{tab:training_strategy}.
\begin{table*}[h]
    \hfill
    \setlength{\tabcolsep}{12pt}
    \renewcommand{\arraystretch}{1.2}
    \begin{tabular}{@{}ll|c|c|c@{}}
    \toprule
    & & \textbf{Pre-Training} & \multicolumn{2}{c}{\textbf{Instruction Tuning}} \\ \cmidrule(l){4-5}
    & & & \textbf{Image-Text}& \textbf{Video} \\
    \midrule 
    \multirow{2}{*}{\rotatebox[origin=c]{90}{\footnotesize \textit{Vision}}}
    & \textbf{Resolution}  & 224 & 224\footnotesize{$\times$\{2$\times$2,~1$\times$\{2,3\},~\{2,3\}$\times$1\}} & 224\footnotesize{$\times$\{2$\times$2,~1$\times$\{2,3\},~\{2,3\}$\times$1\}}  \\
    & \#Tokens & 196 & Max 196\footnotesize{$\times$5} & Max 196\footnotesize{$\times$5} \\
    \midrule 
    \multirow{2}{*}{\rotatebox[origin=c]{90}{\footnotesize \textit{Data}}}
    & \textbf{Dataset} & LCS~\cite{llava-1.5} & LLaVA-Next 779K~\cite{llavanext} & LLaVA-Video-178K subset~\cite{llava-video-178k}
     \\
    & \#Samples & 558K & 779K & 86K  \\
    \midrule
    \multirow{1}{*}{\rotatebox[origin=c]{90}{\footnotesize \textit{Model}}}
    & \textbf{Trainable} & Projector & Projector + LLM & Projector + LLM  \\
    & 7B Vicuna 1.5 LLM & 20.0M & 6.8B & 6.8B  \\
    \midrule 
    \multirow{3}{*}{\rotatebox[origin=c]{90}{\footnotesize \textit{Training}}}
    & \textbf{Batch Size} & 256 & 128 & 128 \\
    & \textbf{LR: $\psi_{\text{vision}}$} & N/A & N/A & N/A \\    
    & \textbf{LR: $\{\theta_{\text{proj}}, \phi_{\text{LLM}}\}$} & 1$\times 10^{-3}$ & 2 $\times 10^{-5}$ & 2 $\times 10^{-5}$  \\
    & \textbf{Epoch} & 1 & 1 & 1  \\
    \bottomrule
    \end{tabular}
    \vspace{1mm}
    \caption{Detailed configuration for each training stage of our LLaVA-NeXT~(StreamFormer).}
    \label{tab:training_strategy}
\end{table*}

\section{More Results}
We add more results of StreamFormer in this section. 

\subsection{Video Action Recognition}
In Table \ref{tab:trecvit}, we add more comparisons with recent causal and autoregressive models, including TRecViT~\cite{trecvit} and Toto~\cite{toto} on Video Action Recognition tasks of K400 and SSv2. While StreamFormer achieves stronger results on K400 and is competitive on SSv2, we emphasize that StreamFormer's key advantage is \textbf{multitask learning across different spatio-temporal granularities}—from short-term action recognition to fine-grained, frame-level tasks like OAD and Online VIS.

\subsection{Downstream Evaluations}
In Table \ref{tab:downstream}, we continue to supplement more results with TVSeries~\cite{tvseries} (w/o flow) for Online Action Detection and OVIS~\cite{ovis} (224p and w/o COCO pre-training) for Online Video Instance Segmentation.In Table \ref{tab:streamingbench}, we detail our results in Real-Time Visual Understanding task of StreamingBench~\cite{streamingbench}. Note that our LLM is Vicuna-7B-v1.5.

\subsection{Qualitative Results}
\begin{figure}[t]
  \centering
    \centering
    \includegraphics[width=\linewidth]{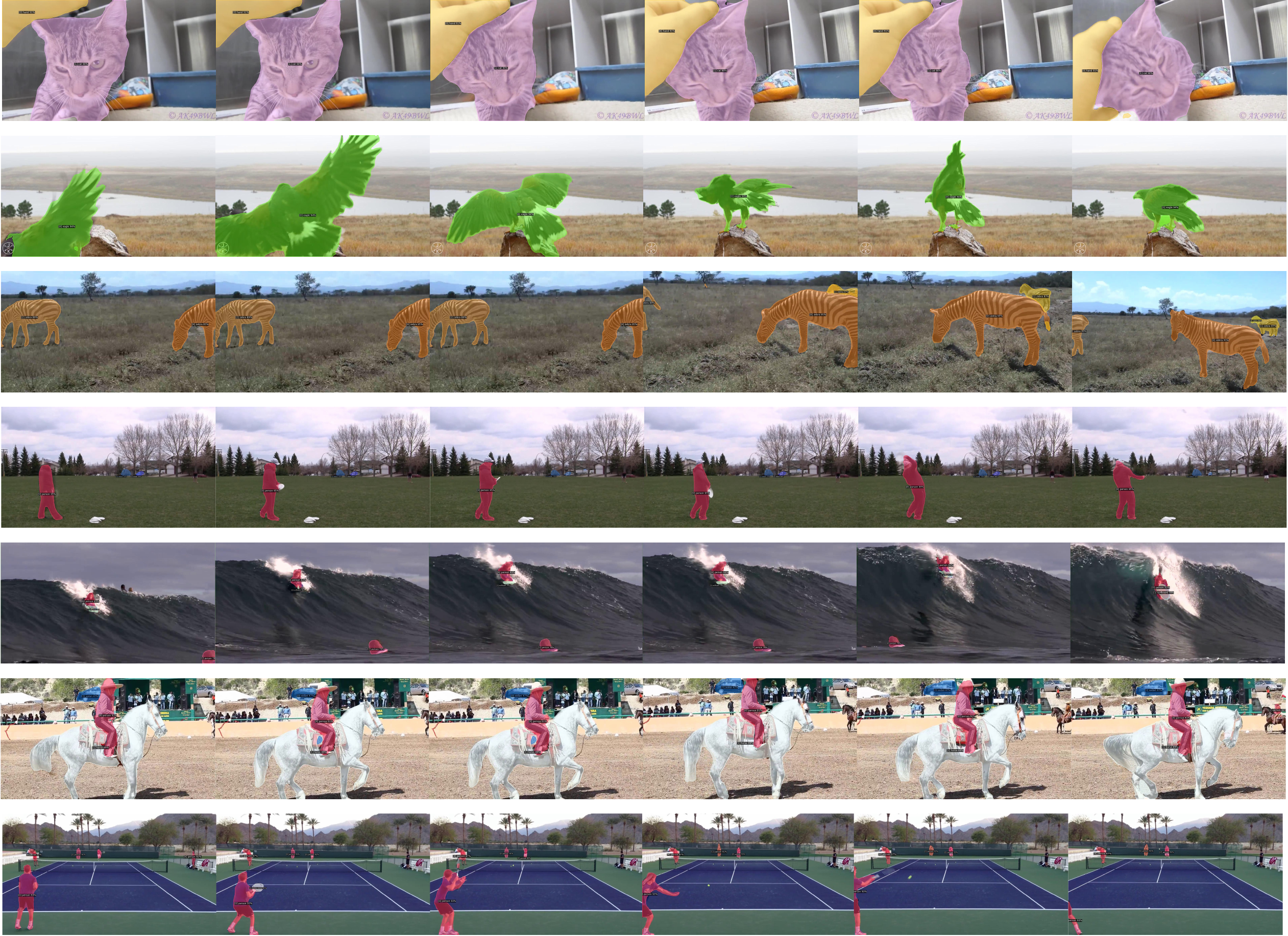}
    \caption{Qualitative results of Online Video Instance Segmentation. Zoom in for a better view.}
    \label{fig:Qualitative}
\end{figure}

We also add qualitative results of StreamFormer's Online Video Instance Segmentation in Figure ~\ref{fig:Qualitative}. Frames are sampled from the validation set of Youtube-VIS 2019.

\begin{table}[h]
\setlength{\tabcolsep}{1 pt}
\parbox{.48\linewidth}{
\centering
\setlength{\tabcolsep}{4pt}
\begin{tabular}{lcc}
\toprule
    Backbone &  K400  &  SSv2 \\
    \midrule
    Toto & 64.7 & - \\
    TRecViT & 78.4 & 66.8 \\
    StreamFormer & 82.2 & 66.5 \\
    \bottomrule
\end{tabular}

\caption{Comparison to other causal backbones.}
\label{tab:trecvit}
}  
\hfill
\parbox{.48\linewidth}{
\centering
\setlength{\tabcolsep}{2pt}
\vspace{-0.4mm}
\begin{tabular}{lcc}
\toprule
    \multirow{2}{*}{Backbone} & TVSeries & OVIS \\
    \cmidrule{2-3}
    & mAP & AP \\
    \midrule
    SigLIP & 84.8 & 18.2 \\
    StreamFormer & 88.1 & 20.8 \\
    \bottomrule
\end{tabular}

\caption{Comparison on additional downstream datasets.}
\label{tab:downstream}
}
\end{table}

\begin{table}[htbp]
\centering

\renewcommand{\arraystretch}{1.4}
\huge
\fontseries{b}
\setlength{\arrayrulewidth}{1pt}
\begin{adjustbox}{max width=\textwidth}
\begin{tabular}{lcccccccccccc}
\hlinewd{2pt}

\multirow{2}{*}{\textbf{Model}} & \multirow{2}{*}{\textbf{Frames}} & \multicolumn{10}{c}{\textbf{StreamingBench Real-Time Visual Understanding}} \\
\cmidrule(lr){3-13}  &  & OP & CR & CS & ATP & EU & TR & PR & SU & ACP & CT & \textbf{All} \\

\hlinewd{1.5pt}
InternVL-V2 & 16 & 68.12 & 60.94 & 69.40 & 77.12 & 67.70 & 62.93 & 59.26 & 53.25 & 54.96 & 56.48 & 63.72 \\
Kangaroo & 64 & 71.12 & 84.38 & 70.66 & 73.20 & 67.08 & 61.68 & 56.48 & 55.69 & 62.04 & 38.86 & 64.60 \\
LongVA & 128 & 70.03 & 63.28 & 61.20 & 70.92 & 62.73 & 59.50 & 61.11 & 53.66 & 54.67 & 34.72 & 59.96 \\
VILA-1.5 & 14 & 53.68 & 49.22 & 70.98 & 56.86 & 53.42 & 53.89 & 54.63 & 48.78 & 50.14 & 17.62 & 52.32 \\
Video-CCAM & 96 & 56.40 & 57.81 & 65.30 & 62.75 & 64.60 & 51.40 & 42.59 & 47.97 & 49.58 & 31.61 & 53.96 \\
Video-LLaMA2 & 32 & 55.86 & 55.47 & 57.41 & 58.17 & 52.80 & 43.61 & 39.81 & 42.68 & 45.61 & 35.23 & 49.52 \\
\hlinewd{1.5pt}
LLaVA-Next (\textbf{StreamFormer}) & 16 & 69.16 & 66.67 & 63.10 & 57.40 & 70.2 & 50.00 & 67.86 & 52.8 & 66.67 & 12.5 & 58.24 \\
\hlinewd{2pt}
\end{tabular}
\end{adjustbox}
\caption{Performance comparison on StreamingBench Real-Time Visual Understanding tasks. The abbreviations are defined as follow: ``OP'' for Object Perception, ``CR'' for Causal Reasoning, ``CS'' for Clips Summarization, ``ATP'' for Attribute Perception, ``EU'' for Event Understanding, ``TR'' for Text-Rich Understanding, ``PR'' for Prospective Reasoning, ``SU'' for Spatial Understanding, ``ACP'' for Action Perception and ``CT'' for Counting.}
\label{tab:streamingbench}
\end{table}

\end{document}